%% file: main.tex
\documentclass[10pt,twocolumn,letterpaper]{article}
\usepackage[accsupp]{axessibility} 
\usepackage{cvpr}              % To produce the CAMERA-READY version
\usepackage{amsmath}
\usepackage{graphicx}
\usepackage{amsmath}
\usepackage{amssymb}
\usepackage{booktabs}
\usepackage[table,xcdraw]{xcolor}
\usepackage{tabularx}
\usepackage{float}
\usepackage{caption}
\usepackage[accsupp]{axessibility}
\graphicspath{{./img/}}

\usepackage[pagebackref,breaklinks,colorlinks]{hyperref}

\usepackage[capitalize]{cleveref}
\crefname{section}{Sec.}{Secs.}
\Crefname{section}{Section}{Sections}
\Crefname{table}{Table}{Tables}
\crefname{table}{Tab.}{Tabs.}

\def\cvprPaperID{10439} % *** Enter the CVPR Paper ID here
\def\confName{CVPR}
\def\confYear{2022}

\usepackage{bibentry}
\usepackage{newfloat}
\usepackage{listings}
\floatstyle{ruled}
\newfloat{listing}{tb}{lst}{}
\floatname{listing}{Listing}

\title{Uni6D: A Unified CNN Framework without Projection Breakdown for 6D Pose Estimation}

\author{Xiaoke Jiang\textsuperscript{1} \qquad Donghai Li\textsuperscript{1}  \qquad Hao Chen\textsuperscript{1}  \qquad Ye Zheng\textsuperscript{2,3} \qquad Rui Zhao\textsuperscript{1,4} \qquad Liwei Wu\textsuperscript{1}\\
SenseTime Research\textsuperscript{1}
Institute of Computing Technology, Chinese Academy of Sciences\textsuperscript{2}\\
University of Chinese Academy of Sciences\textsuperscript{3}\qquad
\\Qing Yuan Research Institute, Shanghai Jiao Tong University, Shanghai, China\textsuperscript{4}\\
{\tt\small \{jiangxiaoke,lidonghai,chenhao2,zhaorui,wuliwei\}@sensetime.com}
\\{\tt\small zhengye@ict.ac.cn}
}

\begin{document}
\maketitle

\input{sec/abs}

\input{sec/intro}
\input{sec/related}
\input{sec/method}

\input{sec/exp_overview}
\input{sec/exp_ablation}

\input{sec/conclusion}

\input{sec/appendix}

{\small
\bibliographystyle{unsrt}
\bibliography{ref}
}
\end{document}

%% file: sec/abs.tex
\begin{abstract}
    As RGB-D sensors become more affordable, using RGB-D images to obtain high-accuracy 6D pose estimation results becomes a better option. State-of-the-art approaches typically use different backbones to extract features for RGB and depth images. They use a 2D CNN for RGB images and a per-pixel point cloud network for depth data, as well as a fusion network for feature fusion. We find that the essential reason for using two independent backbones is the ``projection breakdown'' problem. In the depth image plane, the projected 3D structure of the physical world is preserved by the 1D depth value and its built-in 2D pixel coordinate (UV).
    Any spatial transformation that modifies UV, such as resize, flip, crop, or pooling operations in the CNN pipeline, breaks the binding between the pixel value and UV coordinate. As a consequence, the 3D structure is no longer preserved by a modified depth image or feature.
    To address this issue, we propose a simple yet effective method denoted as \textbf{Uni6D} that explicitly takes the extra UV data along with RGB-D images as input. Our method has a \textbf{Uni}fied CNN framework for \textbf{6D} pose estimation with a single CNN backbone. In particular, the architecture of our method is based on Mask R-CNN with two extra heads, one named RT head for directly predicting 6D pose and the other named abc head for guiding the network to map the visible points to their coordinates in the 3D model as an auxiliary module.
    This end-to-end approach balances simplicity and accuracy, achieving comparable accuracy with state of the arts and 7.2$\times$ faster inference speed on the YCB-Video dataset.
\end{abstract}

%% file: sec/intro.tex
\section{Introduction}

6D pose estimation plays a fundamental role in emerging applications, e.g., autonomous driving~\cite{geiger2012we,xu2018pointfusion,chen2017multi}, intelligent robotic grasping~\cite{collet2011moped,tremblay2018deep, pvn3d} and augmented reality~\cite{marchand2015pose}.
\begin{figure}[t]
  \centering
  \begin{subfigure}{\linewidth}
    \includegraphics[width=\textwidth]{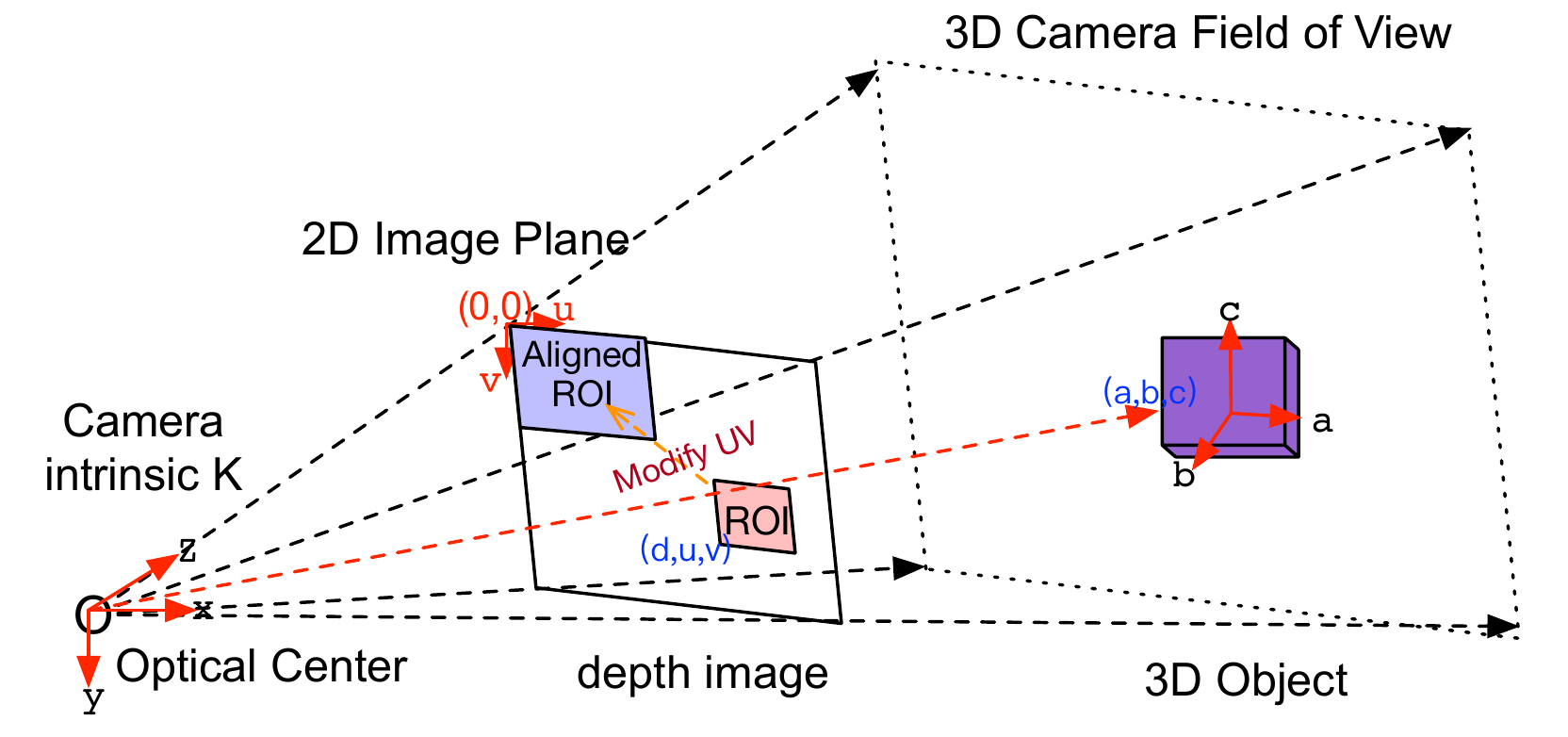}
    \caption{Projection breakdown caused by the RoI transformation, including crop, resize and RoI-Align~\cite{maskrcnn}: The red dotted line connects a 3D object and its projected RoI in the depth image. Any pixel $(d,u,v)$ in the RoI and its corresponding point $(a,b,c)$ on the 3D object follow the projection equation. The equation no longer holds if the built-in coordinate (UV) is modified by RoI-Align, as though the RoI was moved to the top left corner of the image.}
    \label{fig:breakdown-demo}
  \end{subfigure} 
  \hfill
  \begin{subfigure}{\linewidth}
    \includegraphics[width=\textwidth]{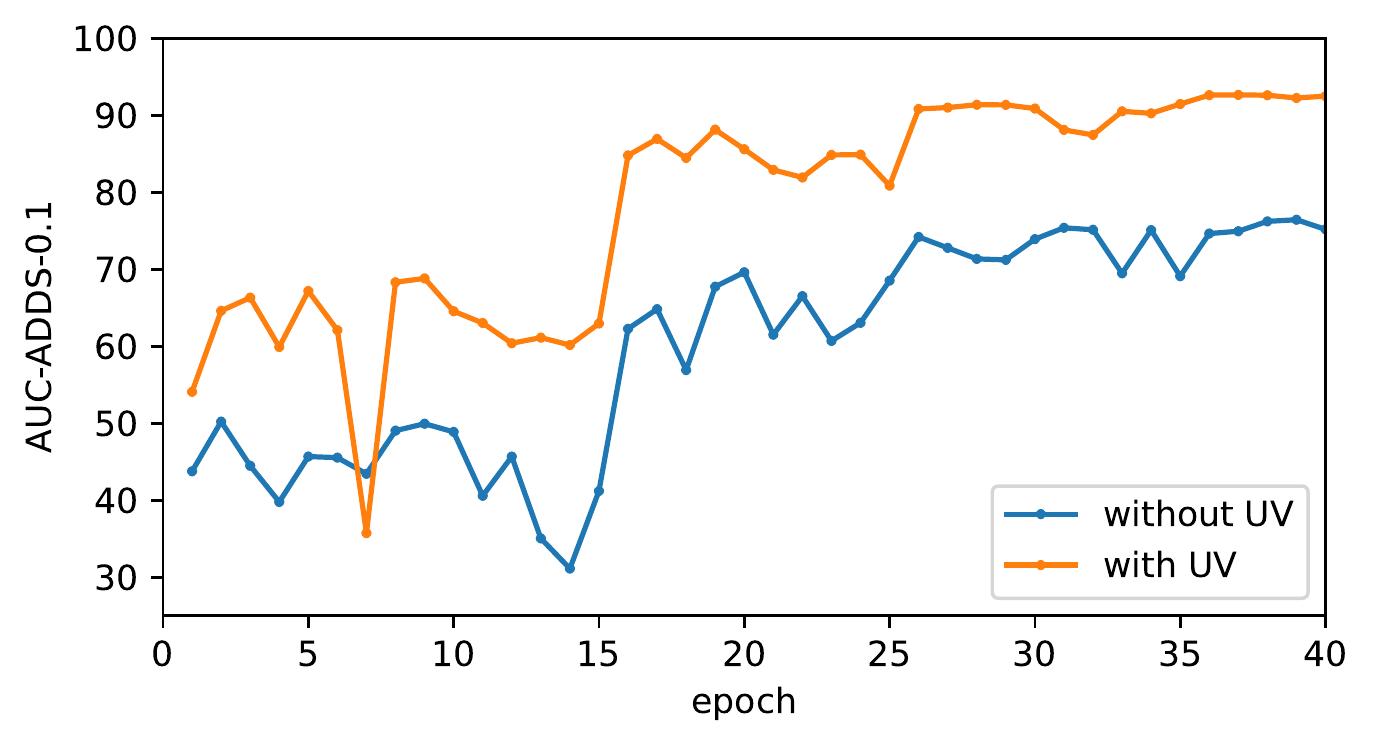}
    \caption{Introducing UV data fixes the problem of projection breakdown and improve the 6D pose estimation performance with various spatial transformations. $x$ axis is the training epoch, and $y$ axis is the testing accuracy.}
    \label{fig:breakdown-safe}
  \end{subfigure}
  \caption{Visualization and experiment results of projection breakdown problem.}
  \label{fig:breakdown}
\end{figure}
RGB-D sensors provide a direct signal of the surface texture and geometry of the physical 3D world, and as their prices fall, RGB-D images are becoming a more attractive data source for 6D pose estimation.

However, state of the arts~\cite{qi2018frustum,wada2020morefusion,densefusion,zhou2020novel,xu2018pointfusion,pereira2020maskedfusion} typically use two separate backbone networks to extract features for RGB and depth images, with a 2D CNN for RGB images and per-pixel PointNet~\cite{pointnet} or PointNet++~\cite{pointnet++} for depth data. After these features have been obtained, an additional fusion process is designed to blend them. Existing methods use different backbones to account for the heterogeneity of these two types of data. However, the key reason they can't extract features with a single backbone is hidden in the classic 3D vision projection equation:
\begin{equation} \label{equ:prj}
  \begin{bmatrix}
    u \\
    v \\
    1 
  \end{bmatrix}
    =
    \frac{K}{d}
  \begin{bmatrix}
    x \\
    y \\
    d 
  \end{bmatrix}
    =
  \frac{K}{R_z(a,b,c)+T_z}
    \begin{pmatrix} 
    R\times \begin{bmatrix}
    a \\
    b \\
    c
  \end{bmatrix} + T\end{pmatrix},
\end{equation}
which describes a point $(a,b,c)$\footnote{We refer a point by its coordinate in object coordinate system.} first rotated by the rotation matrix $R\in SO(3)$ and translated by $T \in R^3$ to the position $(x,y,d)$ in camera coordinate system. It is then projected to the pixel at $(u,v)$ in image plane using the intrinsic matrix $K$ of the camera. The 3D structure of the visible points in the 3D model seen from the camera's perspective is preserved by the depth value ($d$) of each pixel and its coordinate $(u,v)$ in the depth image.
It implies in a single-channel depth image, each pixel value is linked to its built-in coordinates. The plain UV can be used to preserve the 3D structure and changing it will break the 3D projection equation. However, in the traditional CNN pipeline, spatial transformations such as pooling, crop, RoI-Align~\cite{maskrcnn} in the convolution operator, and resize, crop, and flip in data augmentation do change the UV.
As shown in Fig.~\ref{fig:breakdown-demo}, when those spatial transformations are applied to a depth image, the projection equation is broken, and we call this ``projection breakdown''. This is the primary reason why the traditional CNN pipeline struggles to process RGB images and depth data at the same time.

In this paper, we propose a simple yet effective method to save the projection breakdown: explicitly feeding extra UV data along with depth to 2D CNN.
D+UV acts as 3D data, making the value $(d,u,v)$ at each pixel self-complete and decoupling $d$ from its built-in coordinate in the image plane. Given the self-complete information at each pixel, the projection equation still holds after spatial transformations. As a result, we can extract features from RGB-D images using the traditional CNN pipeline, including its data augmentation methods. 
As shown in Fig~\ref{fig:breakdown-safe}, the accuracy of 6D pose estimation is greatly improved with extra UV data.
Extensive experiments show that transformations in data augmentation greatly harm the accuracy without using UV data, but improve the accuracy with UV data is used.

Eq.~\ref{equ:prj} also indicates that the network should map visible points in an RGB-D image to their original coordinates in the 3D model. As far as we know, Existing keypoint-based 6D pose estimation approaches, such as~\cite{pvn3d} and~\cite{ffb6d}, learn the 3D offset from visible points to selected keypoints and produce state-of-the-art accuracy. However, these methods require the iterative voting and regression mechanism as a post-process operation, which accounts for 92.9\% of total frame processing time in FFB6D~\cite{ffb6d} and 79.2\% in PVN3D~\cite{pvn3d} on the multiple-object dataset~\cite{calli2015ycb}. To achieve an accurate, real-time and practical pipeline, we propose Uni6D, an end-to-end 6D pose estimation network based on Mask R-CNN\cite{maskrcnn} that uses a unified backbone to extract feature from RGB-D images. Mask R-CNN performs the object detection and instance segmentation with parallel multi-head networks. On its basis, we add an extra RT head to predict the rotation matrix $R$ and the translation vector $T$ directly, with another abc head to carry out the mapping of visible points, instead of using the time-consuming post-processing used in previous works.
Without any bells and whistles, Uni6D achieves 95.2\% in terms of AUC of ADDS-0.1 on YCB-Video dataset and a real-time inference with 25.6 FPS, which is $7.2\times$ faster than the state of the art.
To summarize, the main contributions of this paper are as follows: (1) We expose the ``projection breakdown'' problem that exists beneath CNN-based depth image processing and introduce the extra UV data into input to fix it, which indicates that a single CNN backbone is all you need for RGB-D feature extraction. (2) We propose an efficient and effective method denoted as Uni6D. The proposed abc head and RT head are optimized in a multitask manner, and we use RT head to directly obtain the pose estimation results. (3) We provide extensive experimental and ablation studies to highlight the benefits of our method, and the results show that our method outperforms existing methods in terms of the time efficiency and achieve promising performance.

%% file: sec/related.tex
\section{Related Work}\label{sec1}
% \begin{table*}
% \centering
% \caption{This table indicates the details comparison results of our method with other state-of-the-arts. Compared with other methods, our approach does not need any complex feature fusion module and post-process operation, which ensures the simplicity and efficiency of it.}
% \label{tab:comp}
%     \begin{tabular}{l|m{2cm}|m{3cm}|m{3cm}|m{3cm}|l}
%     \toprule[2pt]
%       & PoseCNN\cite{posecnn} & DenseFusion\cite{densefusion} & PVN3D\cite{pvn3d} & FFB6D\cite{ffb6d} & Ours \\ \midrule[1pt]
%         Input & RGB & RGB-D & RGB-D & RGB-D & RGB-D, UV \\
%         Backbone & CNN & ResNet, PointNet & ResNet, PointNet++ & ResNet, PointNet++ & ResNet \\
%         Fusion & - & DenseFusion & DenseFusion & Bidirectional Fusion & - \\
%         Granularity & per RoI & per pixel & per pixel & per pixel & per RoI \\
%         Pipeline & end-to-end & seg + predict & predict+regress & predict+regress & end-to-end \\
%         Loss & RT + seg & RT+seg & seg+keypoint+center & seg+keypoint & det+seg+RT+abc \\
%         \toprule[2pt]
%     \end{tabular}
% \end{table*}

\subsection{6DoF Pose Estimation from RGB Images}\label{subsec21}
Holistic methods, keypoint-based approaches and dense correspondence methods are the three types of RGB-only pose estimation methods. Holistic methods~\cite{huttenlocher1993comparing,gu2010discriminative,hinterstoisser2011gradient,xiang2017posecnn,li2018deepim,tulsiani2015viewpoints,su2015render,sundermeyer2018implicit,park2020latentfusion} directly estimate the 6D pose of objects in an RGB image. They use rigid templates to compute the best match pose or directly regress the 6D pose with deep neural networks. However, non-linearity of the rotation space limits generalization ability of DNN based methods.
On the contrary, keypoint-based approaches~\cite{rothganger20063d,kendall2015posenet,oberweger2018making,peng2019pvnet,liu2020keypose} detect the keypoints of objects and then estimate the pose by finding the 2D and 3D correlations at these points. Dense correspondence methods~\cite{glasner2011aware,brachmann2014learning,kehl2016deep,doumanoglou2016recovering,li2019cdpn,wang2019normalized,chen2020learning,hodan2020epos,cai2020reconstruct} utilize the correspondence between image pixels and mesh vertexes to estimation the pose with the Perspective-n-Point (PnP) method to recover poses. Although these dense correspondence method can improve robustness in occlusion situations, the performance is damaged by the unlimited output space. 
 However, the performance of all these methods which only uses RGB images are limited by the loss of geometry information. Therefore, many point clouds based methods have been proposed.
 
\subsection{6DoF Pose Estimation from Point Clouds}\label{subsec22}
 As the cost of depth sensors decreases and accuracy increases, several point clouds based methods has been proposed. The mainstream methods usually use 3D ConvNets~\cite{song2014sliding,song2016deep} or point cloud network~\cite{zhou2018voxelnet,zhou2018voxelnet} to extract the feature and estimate the 6D pose. However, the performance of these methods are limited by the essential noise and shortcomings of the point clouds data. They do not perform well with the sparse and weakly textured point cloud, which indicates the necessity of RGB data.

\subsection{6DoF Pose Estimation from RGB-D Data}\label{subsec23}
Recently, most of 6D pose estimation state-of-the-art approaches utilize the RGB images and depths or point cloud data to achieve a higher accuracy. Classical methods~\cite{hinterstoisser2011multimodal,hinterstoisser2012model,rios2013discriminatively,brachmann2014learning,kehl2016deep,tejani2014latent,wohlhart2015learning} adopt the correspondence grouping and hypothesis verification for the rule based feature and templates of RGB-D data. With the development of deep learning, most data-driven methods use deep neural networks to extract features instead of hand-coded features. They extract features of RGB images and point clouds with independent feature extractors and then fuse them. Most of them~\cite{xu2018pointfusion,qi2018frustum,densefusion,wada2020morefusion,ffb6d} use 2D CNN for RGB images and point cloud network for depth data considering the data heterogeneity between them.
Since the appearance and geometry features are extracted separately, they need to add more complex structure to iterative fuse two features into one in order to predict poses~\cite{qi2018frustum,densefusion,wada2020morefusion,ffb6d} and enable the communication between the RGB and depth information~\cite{ffb6d}. 
\begin{figure*}[h]
    \centering
    \includegraphics[width=0.9\textwidth]{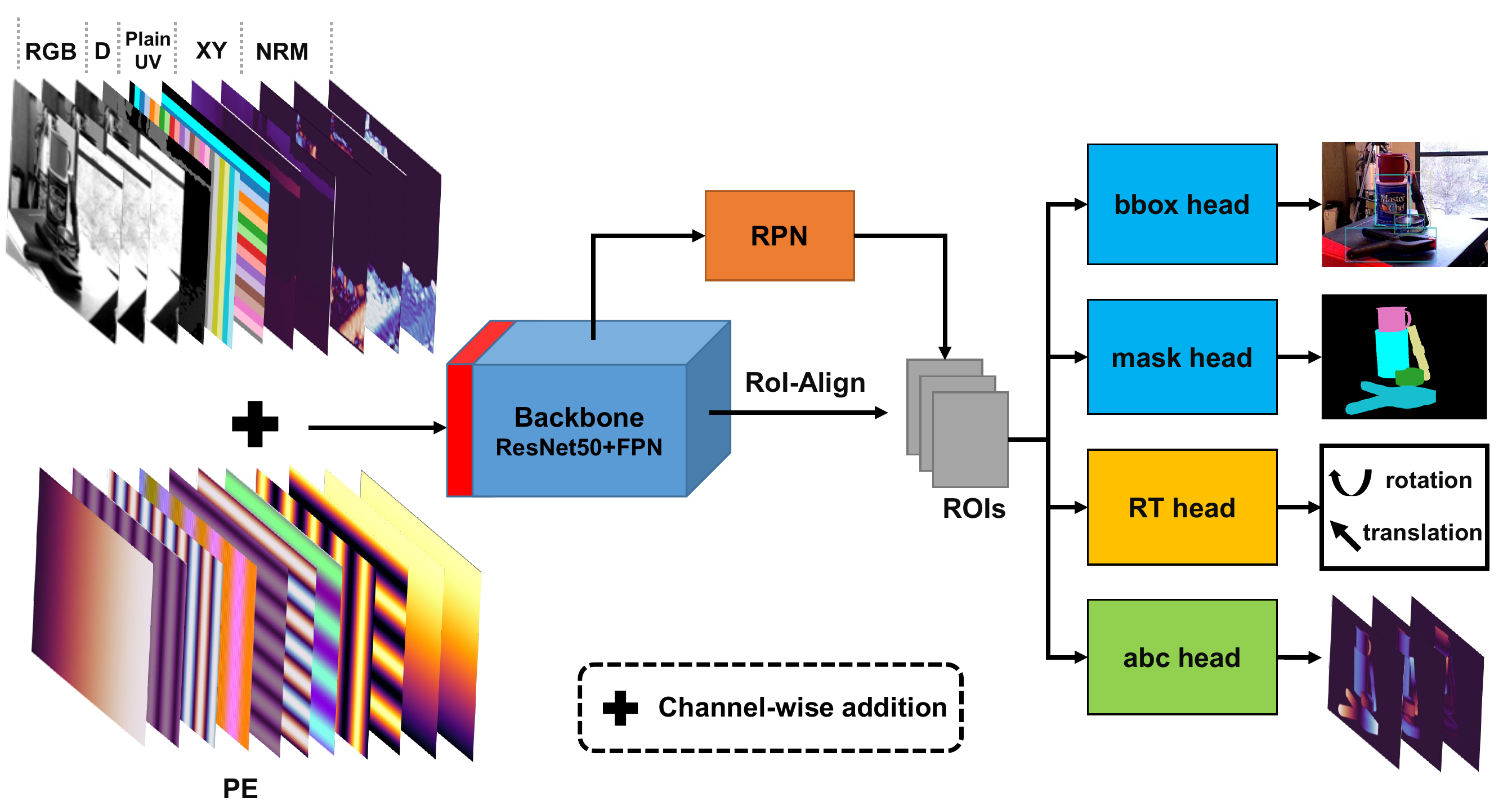}
    \caption{Overview architecture of our Uni6D, which inherits the basic network structure and the simplicity from Mask R-CNN. We add the RGB-D, UV encoding and NRM with PE in a channel-wise manner and feed this new input into the single backbone network. No fusion operation for RGB and depth feature is further required. The red part in the backbone network denotes the channel modification to adapt the RGB-D+UV+NRM (with PE) input, which is to increase the input channel of the first convolutional layer. RT head is used to predict rotation and translation directly and abc head aims to map visible points to their 3D coordinate in its 3D model. With above components, our Uni6D predicts object categories, localization and its RT matrix directly without any post-processing.}
    \label{fig:framework}
\end{figure*}
\subsection{Analysis on State-of-the-art Approaches}
As far as we know, the best result is obtained by two keypoint-based approaches, PVN3D~\cite{pvn3d} and FFB6D~\cite{ffb6d} for now.
As the name implies, keypoint-based approaches design a network to predict the 3D offset from a visible point to some selected keypoints.
Based on this offset, each visible point gives a prediction of keypoints in the camera coordinates system, i.e. $(x,y,z)$.
An iterative voting mechanism is designed\cite{pvn3d} to find the best prediction of keypoints by choosing the clustering with most votes.
The coordinates of the keypoints, i.e. $(a,b,c)$ in 3D model, is known before inference.
Given some matched pairs of $(a,b,c)$ and $(x,y,z)$, an iterative least-square regression is adopted to get the final 6D pose.
We believe that the success of keypoint-based approaches is relies on two reasons:
1) Dense prediction and intelligent de-noising: each visible point will give prediction to visible points and voting is used to filter out bad predictions. 
Given depth images contain a lot of noise, voting is a great way of de-noising.
2) The 3D offset loss is useful to guide the network mapping visible points to its 3D model at pixel level.

PointNet++\cite{pointnet, pointnet++} requires finding K-nearest neighbors in 3D space.
We believe KNN for each pixel, iterative voting and regression are quite heavy when porting to real applications.
We find that the essential reason is the aforementioned \textit{projection breakdown} caused by the change operation on UV. In our work, we reveal and solve the \textit{projection breakdown} problem and adopt a unified feature extractor of the RGB-D data. Benefited from a unified CNN backbone to extract feature from RGB-D images, we propose a straightforward 6D pose estimation pipeline which directly outputs the detection and pose estimation in an end-to-end manner instead of using time-consuming post-processes.
% Some popular and state-of-the-art approaches are  summarized in Table~\ref{tab:comp}.

%% file: sec/method.tex
\section{Methodology of the Uni6D}
Our goal is to propose an accurate, real-time and practical 6D pose estimation method. To this end, we develop an end-to-end framework named Uni6D, which extracts the feature of RGB-D images through a unified backbone network and directly outputs the 6D pose with a straightforward pipeline. Our framework inherits the concise design of Mask R-CNN~\cite{maskrcnn} and introduces a novel RT head and an abc head into it to obtain the 6D pose. With these two heads, we perform multitask joint optimization for the object classification, detection, segmentation and pose estimation. The overall proposed architecture is illustrated in Fig.~\ref{fig:framework}. In this section, we discuss how to inherit the legacy of the Mask R-CNN (Section~\ref{sec:legacy_mask_rcnn}), how to encode UV to fix the projection breakdown (Section~\ref{sec:UV_Input}), the structure of the RT head and abc head (Section~\ref{sec:head}), the loss function (Section~\ref{sec:loss}) and the inference process (Section~\ref{sec:inference}).
\subsection{Legacy of Mask R-CNN} \label{sec:legacy_mask_rcnn}
Object detection and instance segmentation are usually the first steps in 6D pose estimation. As a widely used practical algorithm, Mask R-CNN is capable of achieving them with satisfactory results. Therefore, we use the Mask R-CNN as the basic network. As shown in Fig.~\ref{fig:framework}. Uni6D inherits the basic network structure from Mask R-CNN, including the ResNet~\cite{he2016deep} backbone, the FPN~\cite{lin2017feature} for feature pyramid, the RPN to propose RoIs of potential objects, a mask head for segmentation, and a bbox head for object detection and classification. Using one backbone for two heterogeneous data becomes possible after we solve the projection breakdown problem. As a result, rather than using separate backbone networks for RGB images and depth data as in other methods, we simply use a unified backbone network that can be implemented by making minor changes to Mask R-CNN. Furthermore, feature fusion is no longer required.
Meanwhile, it is very natural to add the regression task of the RT matrix and abc points in parallel on the original basis of Mask R-CNN.  An RT head is proposed to guide the network to map visible points to its 3D model, and an extra abc head is proposed to predict 6D pose directly.

\subsection{Encoding UV Data as Input} \label{sec:UV_Input}
To overcome the issue of projection breakdown, we add the UV data into the input RGB-D data and feed them together into the unified backbone. For the RGB-D data, we directly combine the RGB image and corresponding depth data along the channel dimension. There are three ways to encode positional information. 
1) Plain UV coordinates \textbf{UV}: we directly concatenate this coordinates data with the RGB-D data along the channel dimension. Plain UV is the same height and width with RGB-D images. It has two channels, one channel stores the coordinate of U and the other V, i.e., the value at pixel $(u,v)$ in U-channel is $u$, and in V-channel is $v$.
2) Inverse projected \textbf{XY}: given camera intrinsic matrix $K$ and depth image, we encode corresponding plain UV coordinates $(u,v)$ of each pixel into inverse projected XY $(x,y)$ based on Equation~\ref{equ:prj}. XY has two channels, too. We also concatenate XY with RGB-D data along the channel dimension.
3) Positional encoding \textbf{PE}~\cite{takase2019positional,transformer}: positional encoding is widely used in many other fields such as vision transformers~\cite{transformer}. We encode the position with the trigonometric function and add it into the input data. Overall, the aforementioned three forms of UV encoding information have their own advantages. Plain UV is more direct, XY implies the internal reference information, and PE can be added to other input channels to be more integrated. They work together to complete each other and get the best performance. Since depth normal vector is a widely used for depth image, we also concatenate it to the input and denote it as ``NRM''. 

\subsection{RT head and abc head for Multitask Learning} \label{sec:head}
\begin{figure}[tbp]
    \centering
    \includegraphics[width=\linewidth]{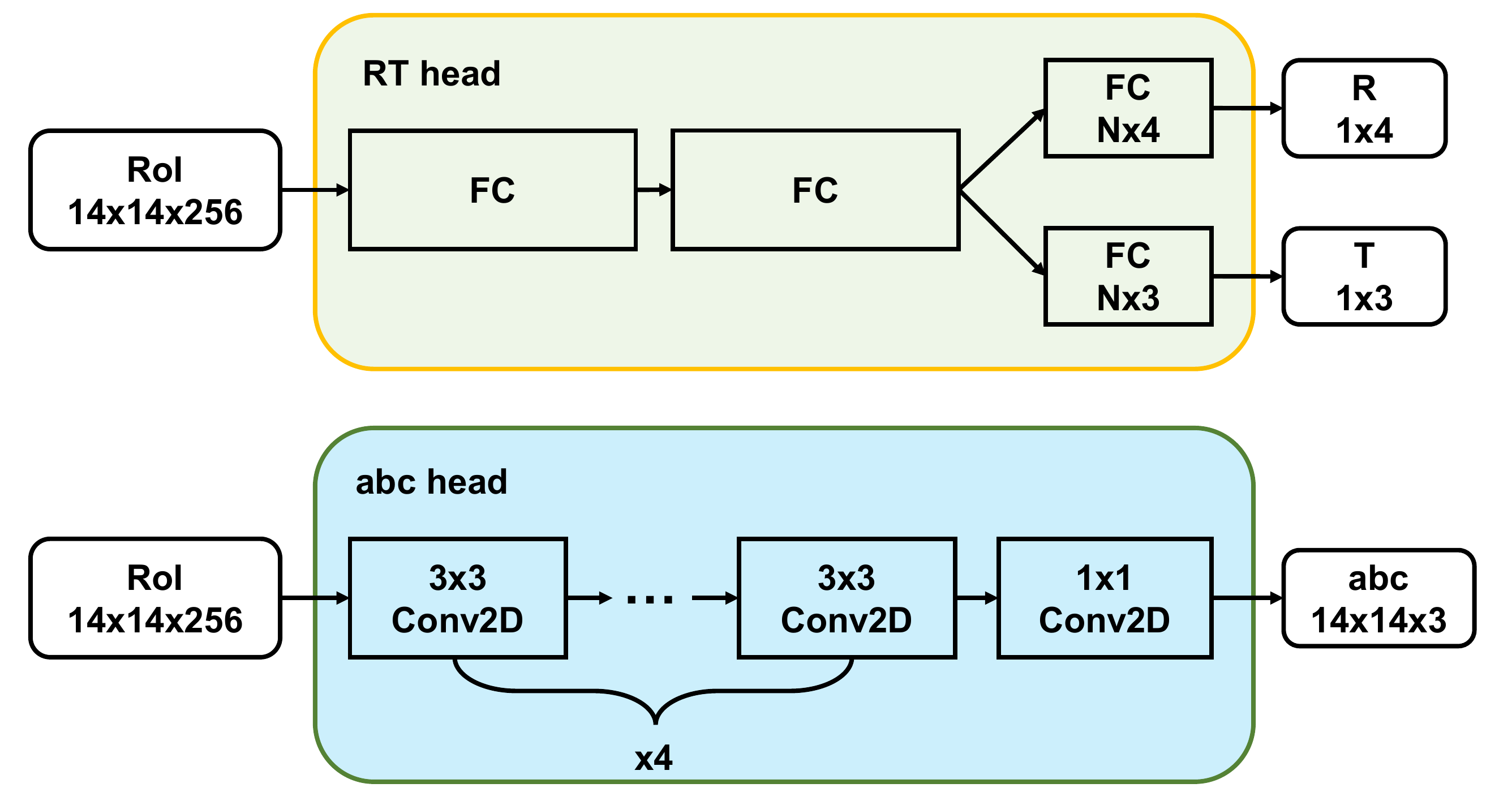}
    \caption{Details of RT head and abc head. Features of all proposals extracted through RoI-Align are used as the input of them. $N$ in the RT head is the number of the input dimension.}
    \label{fig:head}
\end{figure}

We propose RT head and abc head to predict RT matrix and the points of 3D model, where $R$ is the rotation matrix in the form of quaternion and $T \in R^3$ is the translation matrix. These heads are added to Mask R-CNN as two novel parallel branches and take the features of all proposals extracted through RoI-Align as input. As shown in Fig.~\ref{fig:head}, 
the RT head in our method has a similar structure with the bbox head in Mask R-CNN, which contains two shared and two independent fully connected layers to obtain the RT matrix. The abc head is similar with the mask head as an FCN structure, which is developed with four $3\times 3$ and one $1\times1$ convolutional layers to output the 3D points.

\subsection{Loss Function} \label{sec:loss}
In addition to the classification, detection and segmentation loss functions in the original Mask R-CNN, there are two new loss functions for RT head and abc head. The formula of the RT head loss $\mathcal{L}_{rt}$ is:
\begin{equation}
    \begin{split}
        \mathcal{L}_{rt} = \frac{1}{m}\sum_{x\in\mathcal{O}}||(Rx+T) - (R^*x+T^*)||,
    \end{split}
\end{equation}
where $\mathcal{O}$ is the vertex set sampled from the object's 3D model, $R$ is the rotation matrix and $T$ is the translation vector.
The loss of abc head is:
\begin{equation}
    \begin{split}
        \mathcal{L}_{abc} = |a-a^{*}| + |b-b^{*}| + |c-c^{*}|,
    \end{split}
\end{equation}
where $(a, b, c)$ are the coordinate of points.
Thus, the overall loss function is:
\begin{multline}\label{equ:loss}
	\mathcal{L} = \lambda_0 \cdot \mathcal{L}_{rt} + \lambda_1 \cdot \mathcal{L}_{abc} + \lambda_2 \cdot\mathcal{L}_{mask} \\
    + \lambda_3 \cdot  (\mathcal{L}_{bbox} +\mathcal{L}_{cls}) + \lambda_4 \cdot\mathcal{L}_{rpn},
\end{multline}
where $\lambda_0$, $\lambda_1$, $\lambda_2$, $\lambda_3$ and $\lambda_4$ are the weights for each loss.

\subsection{Inference} \label{sec:inference}
Different with other state of the arts of 6D pose estimation, our method does not use any additional time-consuming post-processing and directly outputs the estimation results from the RT head. The straightforward pipeline of our method significantly improves its inference efficiency and practicality. Corresponding quantitative results of estimation accuracy and time efficiency are shown in Section~\ref{sec:quantitative}.

%\subsection{Implementation and Training}
%\note{Experiments setup: 1080Ti, 40 epoch, lr rate, LSBE}
% RoI is aligned to $14\times 14$ by RPN using RoI-Align\cite{maskrcnn}.
% Our model is trained on 16 Nvidia 1080Ti GPUs with 40 epochs in total, and batch size is 3.
% Learning rate is 0.0075 with 4 epochs warm-up. Weight decay (0.1) happens at 15, 25, 35 epoch, respectively.
% The loss balance parameter of RT head, $\lambda 0$, is set to 1 in epoch $[1,20)$, 5 in $[20, 30)$, 20 in $[30, 38)$, 50 in $[38,40]$.

%The increase of $\lambda 0$ with the epoch is to guide the network set RT as the most impo

%% file: sec/exp_overview.tex
\section{Experiments}

\subsection{Benchmark Datasets}
To evaluate our method, we perform experiments on three 6D pose estimation datasets, including YCB-Video, LineMOD and Occlusion LineMOD.

\textbf{YCB-Video}~\cite{calli2015ycb} includes 92 videos with a variety of textures and shapes of RGBD data from 21 YCB objects. All of this data is annotated with 6D poses and instance-level masks. We follow previous work~\cite{posecnn,pvn3d,densefusion} to split the training and testing dataset. We also take the synthetic images for training through the same method in~\cite{posecnn} and apply the hole completion algorithm used in~\cite{pvn3d} for hole filling to the depth images.

\textbf{LineMOD}~\cite{hinterstoisser2011multimodal} contains 13 videos of 13 low-textured objects with the annotation of 6D pose and instance-level mask. We follow previous work~\cite{peng2019pvnet,posecnn} to split the training and testing sets, and we also obtain synthesis images for the training set as the same with~\cite{posecnn,pvn3d}.

\textbf{Occlusion LineMOD}~\cite{brachmann2014learning} is extracted from the LineMOD dataset to evaluate the robustness under heavily occluded situations.
\subsection{Evaluation Metrics}
We evaluate our method following~\cite{posecnn,densefusion,ffb6d} with the average distance metrics ADD and ADD-S. The ADD metric calculates the point-pair average distance between objects vertexes transformed by the predicted pose $[R, T]$ and the ground truth pose $[R^*, T^*]$:
\begin{equation}
    \begin{split}
        \textrm{ADD} = \frac{1}{m}\sum_{x\in\mathcal{O}}||(Rx+T) - (R^*x+T^*)||,
    \end{split}
\end{equation}
where $x$ indicates the vertex sampled from the object's 3D model. ADD-S is applied to symmetric objects based on the closest point distance:
\begin{equation}
    \begin{split}
        \textrm{ADD-S} = \frac{1}{m}\sum_{x_{1}\in\mathcal{O}}\min_{x_{2}\in\mathcal{O}}||(Rx_{1}+T) - (R^*x_{2}+T^*)||.
    \end{split}
\end{equation}
For YCB-Video dataset, we follow~\cite{posecnn,densefusion,pvn3d,ffb6d} to compute the ADD-S and ADD(S) under the accuracy-threshold curve obtained by varying the distance threshold with maximum threshold 0.1 meter (ADD-0.1). For LineMOD dataset, we follow~\cite{peng2019pvnet,ffb6d} to report the accuracy of distance less than 10\% of the objects' diameter (ADD-0.1d).

\begin{table*}[tbp]
\begin{center}
\caption{Evaluation results (\small{ADDS}-S AUC, \small{ADDS}(S) AUC) on the YCB-Video dataset. Symmetric objects are denoted in bold.}
\label{tab:ycb}
\resizebox{0.85\linewidth}{!}{
\begin{tabular}{l|c|c|c|c|c|c|c|c|c|c}
    \toprule[2pt]
 & \multicolumn{2}{c|}{PoseCNN~\cite{posecnn}}  & \multicolumn{2}{c|}{DenseFusion~\cite{densefusion}} & \multicolumn{2}{c|}{PVN3D~\cite{pvn3d}} & \multicolumn{2}{c|}{FFB6D~\cite{ffb6d}} & \multicolumn{2}{c}{Our  Uni6D}   \\ 
    \midrule[1.5pt]
 Object & \small{ADD-S} & \small{ADD(S)} & \small{ADD-S} & \small{\small{ADD(S)}} & \small{ADD-S} & \small{ADD(S)} & \small{ADD-S} & \small{ADD(S)} & \small{ADD-S} &\small{ADD(S)} \\ \midrule[1pt]
        002\_master\_chef\_can & 83.9 & 50.2 & 95.3 & 70.7 & 96 & 80.5 & 96.3 & 80.6 & 95.4 & 70.2 \\
        003\_cracker\_box & 76.9 & 53.1 & 92.5 & 86.9 & 96.1 & 94.8 & 96.3 & 94.6 & 91.8 & 85.2 \\ 
        004\_sugar\_box & 84.2 & 68.4 & 95.1 & 90.8 & 97.4 & 96.3 & 97.6 & 96.6 & 96.4 & 94.5 \\ 
        005\_tomato\_soup\_can & 81.0 & 66.2 & 93.8 & 8.47 & 96.2 & 88.5 & 95.6 & 89.6 & 95.8 & 85.4 \\
        006\_mustard\_bottle & 90.4 & 81.0 & 95.8 & 90.9 & 97.5 & 96.2 & 97.8 & 97.0 & 95.4 & 91.7 \\   % drop
        007\_tuna\_fish\_can & 88.0 & 70.7 & 95.7 & 79.6 & 96.0 & 89.3 & 96.8 & 88.9 & 95.2 & 79.0 \\ 
        008\_pudding\_box & 79.1 & 62.7 & 94.3 & 89.3 & 97.1 & 95.7 & 97.1 & 94.6 & 94.1 & 89.8 \\ 
        009\_gelatin\_box & 87.2 & 75.2 & 97.2 & 95.8 & 97.7 & 96.1 & 98.1 & 96.9 & 97.4 & 96.2 \\  % drop
        010\_potted\_meat\_can & 78.5 & 59.5 & 89.3 & 79.6 & 93.3 & 88.6 & 94.7 & 88.1 & 93.0 & 89.6 \\ 
        011\_banana & 86.0 & 72.3 & 90.0 & 76.7 & 96.6 & 93.7 & 97.2 & 94.9 & 96.4 & 93.0 \\ % drop
        019\_pitcher\_base & 77.0 & 53.3 & 93.6 & 87.1 & 97.4 & 96.5 & 97.6 & 96.9 & 96.2 & 94.2 \\  % drop
        021\_bleach\_cleanser & 71.6 & 50.3 & 94.4 & 87.5 & 96.0 & 93.2 & 96.8 & 94.8 & 95.2 & 91.1 \\  % drop
        \textbf{024\_bowl} & 69.6 & 69.6 & 86.0 & 86.0 & 90.2 & 90.2 & 96.3 & 96.3 & 95.5 & 95.5 \\ 
        025\_mug & 78.2 & 58.5 & 95.3 & 83.8 & 97.6 & 95.4 & 97.3 & 94.2 & 96.6 & 93.0 \\  % drop
        035\_power\_drill & 72.7 & 55.3 & 92.1 & 83.7 & 96.7 & 95.1 & 97.2 & 95.9 & 94.7 & 91.1 \\  % drop
        \textbf{036\_wood\_block} & 64.3 & 64.3 & 89.5 & 89.5 & 90.4 & 90.4 & 92.6 & 92.6 & 94.3 & 94.3 \\ 
        037\_scissors & 56.9 & 35.8 & 90.1 & 77.4 & 96.7 & 92.7 & 97.7 & 95.7 & 87.64 & 79.58 \\  % drop
        040\_large\_marker & 71.7 & 58.3 & 95.1 & 89.1 & 96.7 & 91.8 & 96.6 & 89.1 & 96.66 & 92.76 \\
        \textbf{051\_large\_clamp} & 50.2 & 50.2 & 71.5 & 71.5 & 93.6 & 93.6 & 96.8 & 96.8 & 95.93 & 95.93 \\  % drop
        \textbf{052\_extra\_large\_clamp} & 44.1 & 44.1 & 70.2 & 70.2 & 88.4 & 88.4 & 96.0 & 96.0 & 95.82 & 95.82 \\  % drop
        \textbf{061\_foam\_brick} & 88.0 & 88.0 & 92.2 & 92.2 & 96.8 & 96.8 & 97.3 & 97.3 & 96.1 & 96.1 \\ % drop \midrule[1pt]
        Avg & 75.8 & 59.9 & 91.2 & 82.9 & 95.5 & 91.8 & 96.6 & 92.7 & 95.2 & 88.8 \\   \bottomrule[2pt]
\end{tabular}}
\end{center}                        
\end{table*}

\begin{figure}[tbp]
        \centering
        \includegraphics[width=\linewidth]{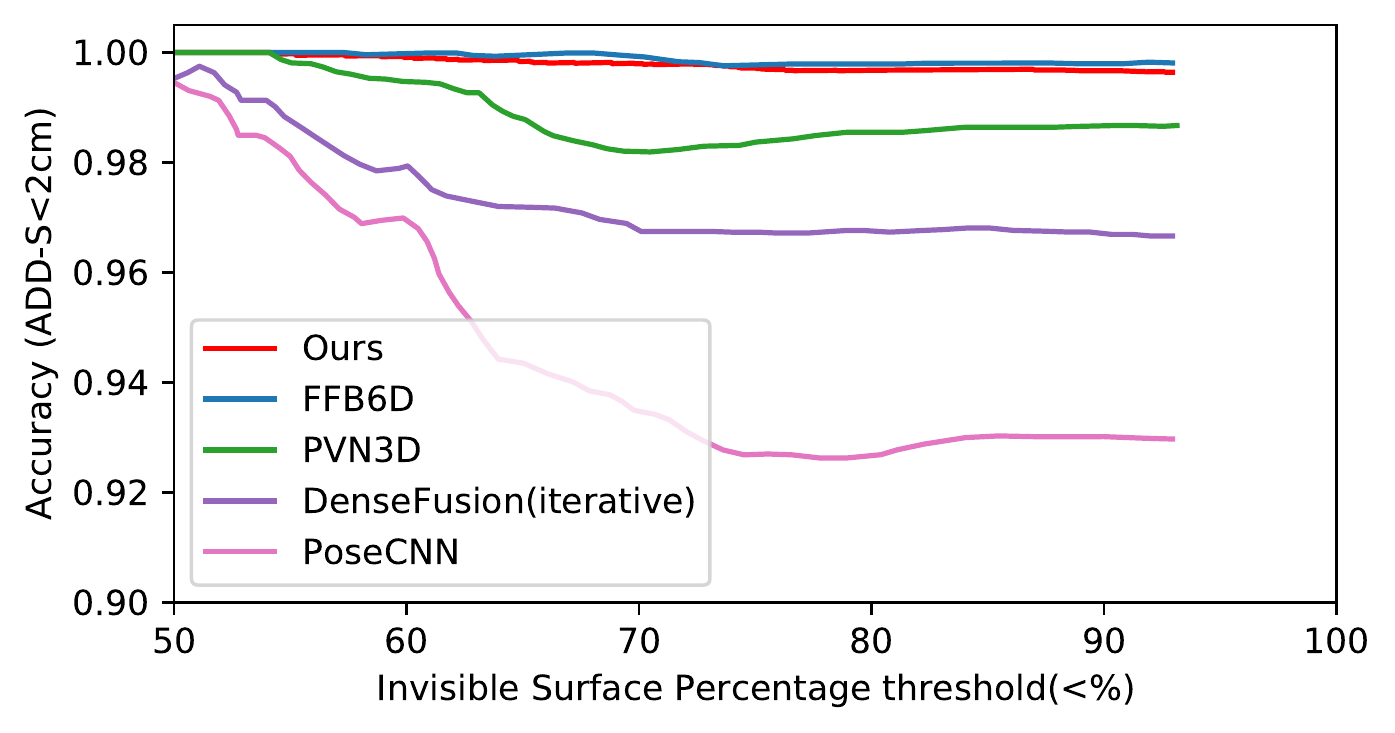}
        \caption{Performance of difference approaches under increasing level of occlusion on YCB-Video Dataset. Our per-RoI method is effective at identifying and handling occlusion.}
        \label{fig:ocYCB}
\end{figure}

\subsection{Quantitative Comparison with Other Methods} \label{sec:quantitative}
We compare the proposed method with others on YCB-Video, LineMOD, and Occlusion LineMOD datasets. Experimental results of YCB-Video dataset are reported in main paper and the results of LineMOD and Occlusion LineMOD are in Appendix.

\textbf{Evaluation results on YCB-Video dataset.} We report the quantitative results of the proposed Uni6D on YCB-Video dataset in Table~\ref{tab:ycb}. In comparison to other methods, our approach achieves 95.2\% on the ADD-S metric and 88.8\% on the ADD(S) metric with a succinct and straightforward pipeline. Although the previous method DenseFusion~\cite{densefusion} does not use post-processing like our method, our method improves the ADD-S by 4\%. In contrast to DenseFusion, which requires two different backbones, CNN and PointNet~\cite{pointnet}, our method only needs a unified CNN to extract RGB-D features. It is worth emphasizing that although the performance of our method is a little bit lower than the current state-of-the-art methods~\cite{pvn3d,ffb6d}, it does not use any additional iterative refinement and post-process operations which are required in state-of-the-art methods. We discuss the time efficiency of our method below. Following~\cite{densefusion, ffb6d}, we also evaluate the robustness towards occlusion in YCB-Video dataset through calculating how the accuracy ($\textrm{ADD-S}<2$cm) changes with extent of occlusion. As indicated in Fig.~\ref{fig:ocYCB}, the performances of our methods has the minimal drop compared with DenseFusion~\cite{densefusion} and FFB6D~\cite{ffb6d}. In particular, the performance under increasing level of occlusion decrease by 0.3\%, which is comparable to FFB6D. Occlusion should be better identified and handled because our RT prediction is based on a RoI.

\begin{table}[tbp]
    \centering
    \caption{Time cost and frames per second (FPS) on YCB-Video Dataset. Our Uni6D is $7.2\times$ faster than FFB6D, and $13.6\times$ than PVN3D. It is capable of estimating 6D pose in real-time. Note that the results of PoseCNN and DenseFusion comes from \cite{densefusion}, which does not state their device type.}
    \label{tab:time-efficiency}
    \resizebox{\linewidth}{!}{
    \begin{tabular}{c|c|c|c|c}
    \toprule[2pt]
        %  PoseCNN+ICP~\cite{posecnn} &  DenseFusion~\cite{densefusion} & Our  Uni6D \\
        Method & Network & Post-process & ALL & FPS \\
        \midrule[1pt]
        PoseCNN+ICP~\cite{posecnn} & 200 & 10400 & 10600 & 0.094 \\
        PoseCNN~\cite{posecnn} & 200 & \textbf{0} & 200 & 5 \\
        DenseFusion~\cite{densefusion} & 50 & 10 & 60 &  16.67\\
        PVN3D~\cite{pvn3d} & 110 & 420 & 530 & 1.89\\
        FFB6D~\cite{ffb6d} & 20 & 260 & 280 & 3.57 \\
        Ours & \textbf{39} & \textbf{0} & \textbf{39} & \textbf{25.64} \\
    \bottomrule[2pt]
    \end{tabular}}
\end{table}

% \textbf{Evaluation results on the LineMOD datasets.} As shown in Table~\ref{tab:lineMOD}, our Uni6D outperforms the other state-of-the-arts with xxx\% ADD-0.1d. 
% For the occlusion LineMOD dataset, we report the results in Table~\ref{tab:oclineMOD}, we can observe that our method achieve xxx \% ADD-0.1d.

\begin{figure}[tbp]
    \centering
    \includegraphics[width=\linewidth]{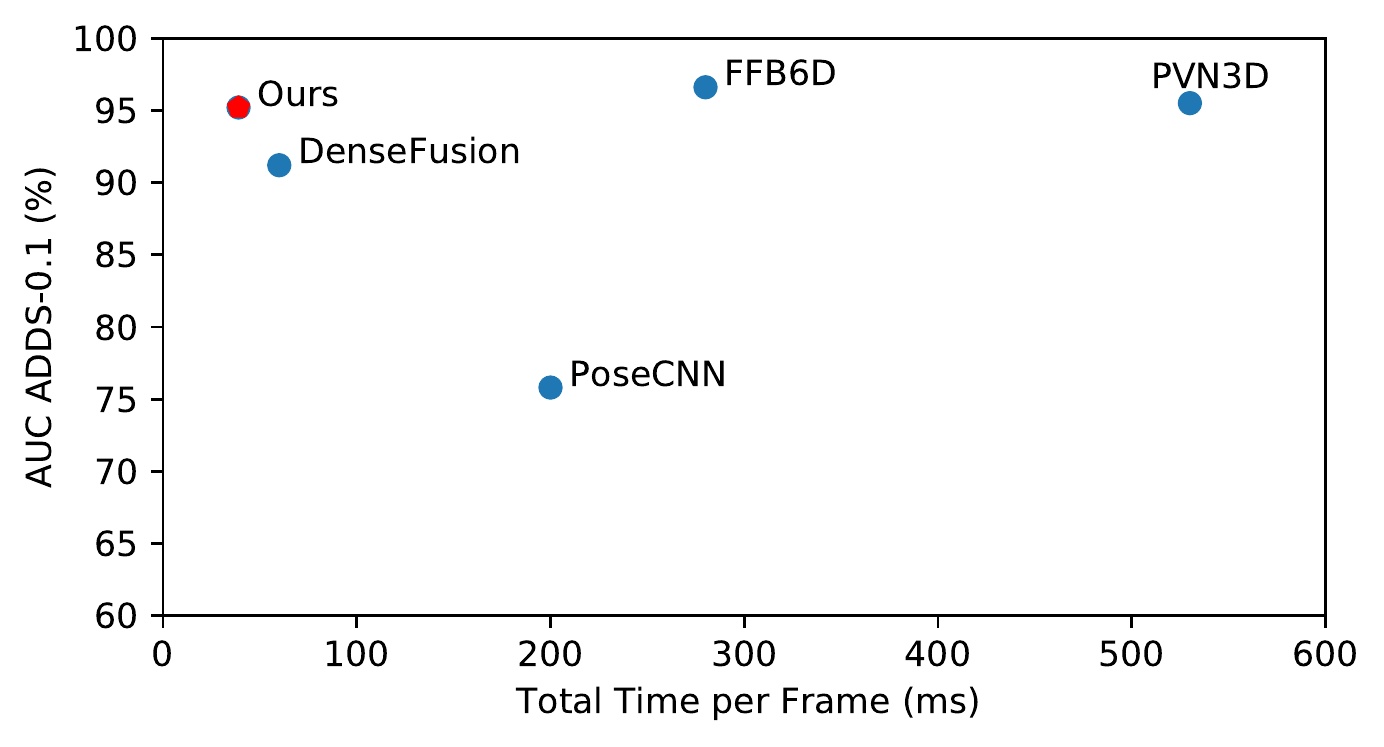}
    \caption{Balance the simplicity/speed and accuracy.}
    \label{fig:acc-time}
\end{figure}

\textbf{Time efficiency.} To highlight the inference efficiency of our straightforward pipeline, we compare the inference speed of our method with PoseCNN+ICP~\cite{posecnn}, PoseCNN~\cite{posecnn}, DenseFusion~\cite{densefusion}, PVN3D~\cite{pvn3d} and FFB6D~\cite{ffb6d} in Table~\ref{tab:time-efficiency}. Post-processing accounts for 92.9\% of the total time in FFB6D and 79.2\% in PVN3D. DenseFusion has a faster inference speed compares to them, while our method further speeds up the inference, benefited from our straightforward pipeline which does not need any post-processing. Our Uni6D is $7.2\times$ faster than FFB6D, and $13.6\times$ than PVN3D. Comparison results of inference time and performance are shown in Fig~\ref{fig:acc-time}.
\subsection{Implementation Details}
In all experiments, we adopt ResNet-50~\cite{he2016deep} as the backbone network with FPN~\cite{lin2017feature}. We train all models with 16 GPUs (three images per GPU) for 40 epochs with an SGD optimizer which momentum is 0.9 and weight-decay is 0.0001. The initial learning rate is set to 0.0075 with a linear warm-up, and decreased by 0.1 after 15, 25 and 35 epochs. The weight hyperparameter $\lambda_0$ in the loss function of RT head  is set to 1 in epoch $[1,20)$, 5 in $[20, 30)$, 20 in $[30, 38)$ and 50 in $[38,40]$.

%% file: sec/exp_ablation.tex
\subsection{Ablation Study}
\subsubsection{Projection Breakdown Saved by UV}
%\textbf{Effects of the positional encoding under varying spatial transformations.}

To investigate the effects of the UV encoding information in the projection breakdown problem, we perform experiments with several spatial transformations on the YCB-Video dataset in Table~\ref{tab:pe}. Using any of the common spatial transformations as training augmentation methods, such as random resize, crop, horizontal flip, and vertical flip, significantly reduces performance by up to 15\% ADD-S and 20\% ADD(S). These results directly reflect the destructiveness of projection breakdown. We add explicit UV encoding information with UV, XY, and PE to solve this issue and make the \textit{one backbone is all you need} possible. 
% \textcolor{red}{In addition, the UV encoding can also improve the performance by 1.2\% ADD(S) in the vanilla CNN pipeline.}

\subsubsection{UV Encoding Methods}
%\subsection{Component-wise Analysis}
%\textbf{UV encoding information}
We investigate the contributions of all components in UV encoding of our method on the YCB-Video dataset. The results are shown in Table~\ref{tab:component-wise}. ``RGB-D" denotes that we only use RGB-D data as input, which is used as the baseline. ``Plain UV" means the plain UV coordinates data. ``XY" represents adding the inverse projected XY coordinates data. ``NRM" means the depth normal vector. ``PE" is the position encoding. ``abc head" denotes adding the abc head during training. We can see that the flip transformation degrades performance in the baseline, and that adding position information to the RGB-D data in the form of plain UV, XY, NRM, or PE can alleviate this degradation. Compared with the baseline, our method brings 4.19\% improvement in ADD-S and 9.11\% in ADD(S). 

% \begin{table}[tbp]
%     \centering
%     \caption{Effects of the UV encoding under varying spatial transformations. Spatial transformations greatly harm the accuracy when no UV encoding information is provided, but improve the accuracy with UV encoding information.}
%     \label{tab:pe}
%     % \begin{tabular}{m{0.8cm}|m{0.6cm}|m{0.6cm}|m{0.6cm}|m{0.6cm}|c|c}
%     \resizebox{\linewidth}{!}{
%     \begin{tabular}{c|c|c|c|c|c|c}
%     \toprule[2pt]
%          UV Enc & Resize & Crop & Hflip & Vflip & ADDS & ADD(S) \\
%         \midrule[1pt]
%           w/o & \checkmark & \checkmark & \checkmark & \checkmark & 76.42 & 62.56  \\
%           w & \checkmark & \checkmark & \checkmark & \checkmark & 92.64 & 82.58 \\
%           \hline
%           w/o & \checkmark & \checkmark & \checkmark &           &  79.45 &65.62  \\
%           w & \checkmark & \checkmark & \checkmark &&   93.22 & 83.99\\
%           \hline 
%           w/o & \checkmark & \checkmark &  & \checkmark &    87.96 & 74.27  \\
%           w & \checkmark & \checkmark &  & \checkmark &   92.79 & 83.47 \\
%           \hline
%           w/o & \checkmark &   &  &   &  90.67 & 79.78\\
%           w & \checkmark &   &  &   &92.65  & 82.91\\
%           \hline   
%           w/o & & \checkmark  &  &   & 91.88 & 81.99 \\
%           w & & \checkmark  &  &  & 92.52 & 84.33\\
%           \hline
%         %   w/o &\checkmark & \checkmark & \checkmark & \checkmark & 76.42 & 61.20  \\
%           w/o & & & & & 92.08 & 82.95  \\
%           w & & & & & 92.63 & 84.13 \\
%     \bottomrule[2pt]
%     \end{tabular}}
% \end{table}

\begin{table}[tbp]
    \centering
    \caption{Effects of the UV encoding under varying spatial transformations. Spatial transformations greatly harm the accuracy when no UV encoding information is provided, but improve the accuracy with UV encoding information.}
    \label{tab:pe}
    % \begin{tabular}{m{0.8cm}|m{0.6cm}|m{0.6cm}|m{0.6cm}|m{0.6cm}|c|c}
    \resizebox{\linewidth}{!}{
    \begin{tabular}{c|c|c|c|c|c|c}
    \toprule[2pt]
         UV Enc & Resize & Crop & Hflip & Vflip & ADD-S & ADD(S) \\
        \midrule[1pt]
          w/o & \checkmark & \checkmark & \checkmark & \checkmark & 78.87 & 65.88  \\
          w & \checkmark & \checkmark & \checkmark & \checkmark & 92.82 & 81.95 \\
          \hline
           w/o & \checkmark & \checkmark & \checkmark &           &  79.01 & 65.79  \\
           w & \checkmark & \checkmark & \checkmark &&   93.61 & 84.36\\
          \hline 
          w/o & \checkmark & \checkmark &  & \checkmark &    90.40 & 79.26 \\
          w & \checkmark & \checkmark &  & \checkmark &   93.18 & 84.31 \\
          \hline
           w/o & \checkmark &   &  &   &  92.92 & 83.39\\
          w & \checkmark &   &  &   & 93.89 & 85.76\\
          \hline   
           w/o & & \checkmark  &  &   & 92.45 & 83.18 \\
          w & & \checkmark  &  &  & 92.96  & 85.05\\
          \hline
        %   w/o &\checkmark & \checkmark & \checkmark & \checkmark & 76.42 & 61.20  \\
        %   w/o & & & & & 92.95 & 84.47  \\
        %   w & & & & & 92.24 & 83.30 \\
    \bottomrule[2pt]
    \end{tabular}}
\end{table}
% % ------------------------------------------------------------------------------------
\subsection{Qualitative Results}
For intuitively comparing the qualitative results between our Uni6D and other methods~\cite{pvn3d,ffb6d} on the YCB-Video dataset, we give some estimation results in Fig.~\ref{fig:vis}. Our Uni6D significantly outperforms the state of the art and performs the most robust in different occlusion situations.\footnote{Results of PVN3D and FFB6D are from the paper of FFB6D~\cite{ffb6d}.}

%% file: sec/conclusion.tex
\begin{figure*}[t]
    \centering
    \includegraphics[width=0.95\textwidth]{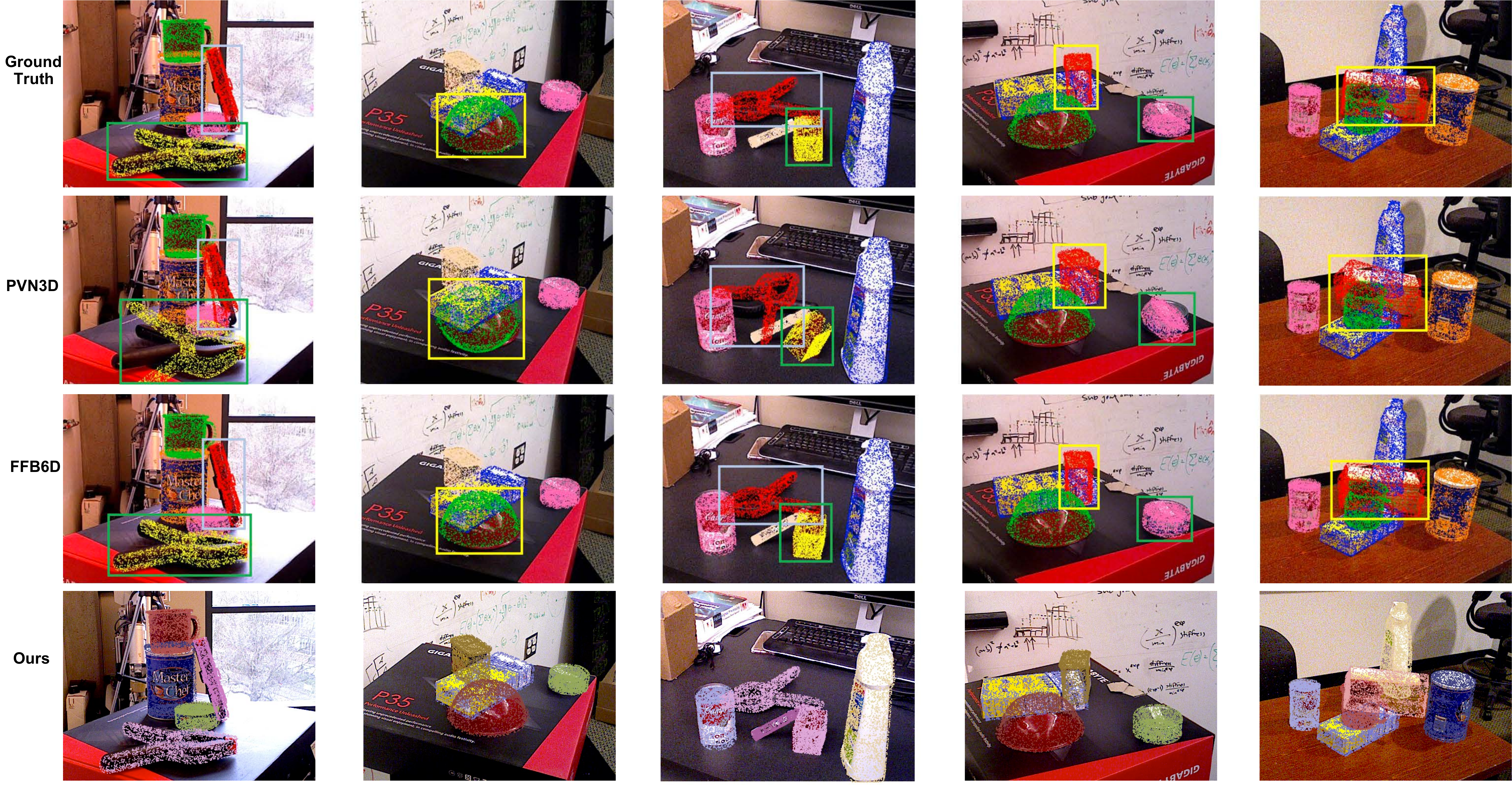}
    \caption{Qualitative results of 6D pose on the YCB-Video dataset.}
    \label{fig:vis}
\end{figure*}
\section{Limitation Analysis}
In this section, we discuss the limitations of our method.

\textbf{For the performance of 6D pose estimation}, our method still has a little gap compared with the state of the art in YCB dataset. The main reason is that we do not use any time-consuming post-processing or iteration refinement in the pipeline of our method. We only add a lightweight abc head with the abc regression task to introduce an auxiliary loss for auxiliary model training. In inference, the abc head is removed and we directly obtain the 6D pose estimation results from the RT head. This straightforward inference pipeline significantly improves the efficiency of inference and reduces the difficulty of engineering implementation. It is worth exploring more efficient post-processing algorithms for our unified 2D CNN to further improve the performance while maintaining real-time inference.
% the training and testing of our end-to-end method is applied in a RoI-wise manner, different with the per-pixel dense prediction widely used in previous works~\cite{posecnn,densefusion,ffb6d}. The advantage of the dense prediction is it can filter out noise signal from noisy depth image to improve the accuracy. We also propose abc head and apply it on each pixel of each \emph{RoI} as an auxiliary task. Moreover, compared to representing a RoI through a high resolution with hundreds of pixels in length and width, the current $(14, 14)$ is still too small. More work should be explored to de-noise RoI features for better result without loss of simplicity.

\textbf{For the RoI-Align operation}, it limits the performance of RT head and reduces the accuracy of our method. The training and testing of our end-to-end method is applied in a RoI-wise manner with a rough object-level feature for each object, which is different from the per-pixel dense prediction widely used in previous works~\cite{posecnn,densefusion,ffb6d}. More research into de-noising RoI features for better results without sacrificing simplicity is needed.

% TODO-reopen ???
% \textbf{The UV encoding method is still naive}, we directly fuse the UV encoding data, includes plain UV, XY and PE with the RGB-D image, more refined and suitable methods need to be explored.

% ------------------------------------------------------------------------------------
% \begin{table}[tbp]
% \begin{center}
% \caption{Effects of each component in our work. Results of ADD(S) AUC and ADD-S AUC on YCB-Video are reported respectively. Compared with the baseline, our method brings 5.08\% improvement in ADD-S and 10.19\% in ADD(S).}
% \label{tab:component-wise}
% \resizebox{1.0\linewidth}{!}{
% \begin{tabular}{cccccc|cc}
% \toprule[2pt]
% RGB-D & Plain UV & XY & PE & NRM & abc head & ADDS & ADD(S) \\
% \midrule[1pt]
% %w flip &  &  &  & &  & 76.21 & 63.13 \\
% \checkmark &  &  &  & &  & 90.14 & 78.41\\
% \checkmark & \checkmark &  &  & &  & 92.26 & 82.29 \\
% \checkmark & ~ & \checkmark &  & &  & 91.42 & 80.69 \\
% \checkmark & ~ & ~ & \checkmark & &  & 92.88 & 83.92 \\
% \checkmark & ~ & ~ & ~ & \checkmark &  & 91.10 & 81.13 \\
% \checkmark & \checkmark & ~ & \checkmark & &  & 93.52 & 84.21\\
% \checkmark & \checkmark & \checkmark & \checkmark &  &  & 93.26 & 84.36 \\
% \checkmark & \checkmark & \checkmark & \checkmark & \checkmark &  & 95.09 & 86.90 \\
% \checkmark & \checkmark & \checkmark & \checkmark & \checkmark & \checkmark & \textbf{95.22} & \textbf{88.60} \\
% \bottomrule[2pt]
% \end{tabular}
% }
% \end{center}
% \end{table}
\begin{table}[tbp]
\begin{center}
\caption{Effects of each component in our work. Results of ADD(S) AUC and ADD-S AUC on YCB-Video are reported respectively. Compared with the baseline, our method brings 4.19\% improvement in ADD-S and 9.11\% in ADD(S).}
\label{tab:component-wise}
\resizebox{1.0\linewidth}{!}{
\begin{tabular}{cccccc|cc}
\toprule[2pt]
RGB-D & Plain UV & XY & PE & NRM & abc head & ADD-S & ADD(S) \\
\midrule[1pt]
%w flip &  &  &  & &  & 76.21 & 63.13 \\
\checkmark &  &  &  & &  & 90.99 & 79.72\\
\checkmark & \checkmark &  &  & &  & 94.06 & 85.39 \\
\checkmark & ~ & \checkmark &  & &  & 94.17 & 85.66 \\
\checkmark & ~ & ~ & \checkmark & &  & 93.54 & 85.05 \\
\checkmark & ~ & ~ & ~ & \checkmark &  & 93.79 & 84.79 \\
\checkmark & \checkmark & \checkmark &  & &  & 93.90 & 85.06\\
\checkmark & & \checkmark & \checkmark  & &  & 93.27 & 84.49\\
\checkmark & \checkmark & ~ & \checkmark & &  & 93.65 & 84.51\\
\checkmark & ~ & \checkmark & ~ & \checkmark &  & 94.70 & 86.76\\
\checkmark & \checkmark & ~ & ~ & \checkmark & & 94.26 & 85.52 \\
\checkmark & ~ & ~ & \checkmark & \checkmark & & 93.31 & 83.42 \\
\checkmark & \checkmark & \checkmark & \checkmark &  &  & 93.55 & 84.83 \\
\checkmark & \checkmark & \checkmark & \checkmark & \checkmark &  & 94.91 & 86.93 \\
\checkmark & \checkmark & \checkmark & \checkmark & \checkmark & \checkmark & \textbf{95.18} & \textbf{88.83} \\
\bottomrule[2pt]
\end{tabular}
}
\end{center}
\end{table}
%-------------------------------------------------------------------------

\section{Conclusions}
In this paper, we reveal the ``projection breakdown'' hidden underneath CNN-based depth image processing, which
explains the two-backbone design for RGB-D processing and the per-pixel feature extraction for depth processing adopted by most of the existing literature. Rather than using two separate backbones, we fix the projection breakdown by explicitly feeding extra UV data along with depth to the backbone. As a result, all you need to extract the feature from RGB-D images is a general 2D CNN backbone.
Thus, we propose Uni6D, an end-to-end 6D pose estimation framework based on Mask R-CNN. Uni6D uses ResNet and FPN as its backbone to extract features from RGB-D images, extra feature fusion is not required anymore. The 6D pose estimation results are obtained directly from the RT head using a straightforward inference pipeline that does not require any time-consuming post-processing. We also develop the abc head as an auxiliary task for training the network in our framework. Extensive experimental results show our method outperforms other state-of-the-art methods in terms of time efficiency, performance, and robustness in the challenging YCB-Video. Worthwhile and related future work can be spawned from the proposed new 6D pose estimation paradigm, which performs 6D pose estimation with a unified CNN framework with the above contributions.
% Uni6D utilizes mask-head of existing Mask R-CNN for instance segmentation and adds an extra head to predict RT directly. 
% We also highlight the rationality of keypoint-based approaches.
% The network is supposed to map visible points on RGB-D image to its 3D model, and keypoint offset loss acts as the role to do the mapping.
% But obtaining final RT from keypoint requires iterative voting and regression, which is time-consuming.
% Uni6D does not predict keypoints to keep simplicity.
% To balance the simplicity and accuracy,
% we add an abc head in multitask fashion to carry out the mapping for visible points, but let the RT head give the final estimation directly. 
%This multi-task design is still end-to-end and helps to obtain an accurate 6D pose estimation.
%Our end-to-end pose estimation reaches comparable results with methods, but 

% One of our future work is to catch state-of-the-art in terms of accuracy, while still keeping our simplicity and speed.
\section*{Acknowledgement}
This work was also sponsored by Hetao Shenzhen-Hong Kong Science and Technology Innovation Cooperation Zone: HZQB-KCZYZ-2021045.

%% file: sec/appendix.tex
\clearpage
\twocolumn[
\begin{@twocolumnfalse}
	\section*{\centering{Supplementary Material for \\ \emph{Uni6D: A Unified CNN Framework without Projection Breakdown for 6D Pose Estimation\\[25pt]}}}
\end{@twocolumnfalse}
]

\section{More implementation details }
\subsection{The details of the positional encoding.}
PE is implemented using equation~\ref{eq1} and the details will be added in the final version.
\begin{equation}\label{eq1}
\begin{aligned}
    & PE(x,y,2i) = sin(x/10000^{(4i/D)}) \\
    & PE(x,y,2i+1) = cos(x/10000^{(4i/D)}) \\
    & PE(x,y,2j+D/2) = sin(y/10000^{(4j/D)}) \\
    & PE(x,y,2j+1+D/2) = cos(y/10000^{(4j/D)}), \\
\end{aligned}
\end{equation}
where $(x,y)$ is a point in 2d space, $i,j$ is an integer in $[0, D/4)$, $D$ is the size of the channel dimension.
\subsection{The details of the pre-trained weight.} 
We use the ImageNet pre-trained weight, and the first convolutional layer is initialized with the kaiming uniform.
\textbf{For YCB dataset}:
\begin{itemize}
    \item Backbone: ResNet50 + FPN;
    \item Input data: RGB-D+UV+PE+XY+NRM, rotation matrices are represented by quaternions, other settings are same with PVN3D~\cite{pvn3d};
    \item Data augmentation: 
    \begin{enumerate}
        \item multi-scale training: [320, 400, 480, 600, 720] (max size is 900);
        \item background replacing: replace the background of the rendered data with the real image background;
        \item random crop: 0.3 probability, need to keep all objects;
    \end{enumerate}
    \item Training:
    \begin{enumerate}
        \item Pretrained: ImageNet;
        \item Schedule: 40epoch, MultiStepLR with [15, 25, 35] schedule and 0.1$\times$ decay ;
        \item Optimizer: SGD, momentum 0.9, weight\_deacy 0.0001, warm-up 4 epoch;
    \end{enumerate}
    \item Loss function:
    \begin{enumerate}
        \item $\mathcal{L} = \lambda_0 \cdot \mathcal{L}_{rt} +  \mathcal{L}_{abc} + \mathcal{L}_{mask} 
    +   \mathcal{L}_{bbox} +\mathcal{L}_{cls} +\mathcal{L}_{rpn} $
        \item  $\lambda_0$ is changed in training: 1-15 epoch is 1, 16-25 epoch is 5, 26-35 epoch is 10 and 36-40 epoch is 20;
    \end{enumerate}
\end{itemize}

\textbf{For Linemode dataset}:
\begin{itemize}
    \item Backbone: ResNet50 + FPN;
    \item Input data: RGB-D+UV+PE+XY+NRM, rotation matrices are represented by quaternions, other settings are same with PVN3D~\cite{pvn3d}, except using camera intrinsics for real data to render data;
    \item Data augmentation: 
    \begin{enumerate}
        \item multi-scale training: [320, 400, 480, 600, 720] (max size is 900);
        \item background replacing: replace the background of the rendered data with the real image background;
        \item random crop: 0.3 probability, need to keep all objects;
        \item random erase: 0.1 probability
    \end{enumerate}
    \item Training:
    \begin{enumerate}
        \item Pretrained: ImageNet;
        \item Schedule: 40epoch, MultiStepLR with [15, 25, 35] schedule and 0.1$\times$ decay ;
        \item Optimizer: SGD, momentum 0.9, weight\_deacy 0.0001, warm-up 4 epoch;
    \end{enumerate}
    \item Loss function:
    \begin{enumerate}
        \item $\mathcal{L} = \lambda_0 \cdot \mathcal{L}_{rt} +  \mathcal{L}_{abc} + \mathcal{L}_{mask} 
    +   \mathcal{L}_{bbox} +\mathcal{L}_{cls} +\mathcal{L}_{rpn} $
        %\item $Loss = \alpha Add-loss + RPN-loss + bbox-loss + cls-loss + mask-loss + abc-loss$;
        \item  $\lambda_0$ is changed in training: 1-15 epoch is 1, 16-25 epoch is 5, 26-35 epoch is 10 and 36-40 epoch is 20;
    \end{enumerate}
\end{itemize}

\section{Ablation Studies of abc Head}
We provide results of more ablation studies for abc head on YCB dataset in Table~\ref{tab:abc}. We combine the abc head with different UV input information to verify the effectiveness of it. We can observe that our abc head can improve the performance \textbf{without} UV and it can further improve the performance \textbf{with} different types of UV. These results demonstrate the effectiveness of abc head as an auxiliary training task.

\begin{table}[htbp]
    \centering
    \resizebox{\linewidth}{!}{
    \begin{tabular}{c|c|c|c|c|c}
    \toprule
        ~ & RGB-D & Plain UV & XY & PE & NRM\\
        \hline
        w/o & 90.99/79.72 & 94.06/85.39 & 94.17/85.66 & \textbf{93.54}/85.05 & 93.79/84.79 \\
        w & \textbf{91.13}/\textbf{80.89} & \textbf{94.49}/\textbf{86.46} & \textbf{94.33}/\textbf{86.90} & 93.53/\textbf{86.09}& \textbf{93.89}/\textbf{84.96}\\
        \bottomrule
    \end{tabular}}
    \caption{Ablation study results of abc head, the format is ADDS/ADD(S).}
    \label{tab:abc}
\end{table}

\section{Quantitative Results on the LineMOD Dataset} 
Experimental results of LineMOD dataset are reported in Table~\ref{tab:LineMOD-ADD}, our approach achieves 97.03\% ADD-0.1d ACC with a succinct and straightforward pipeline compared with other methods. LineMOD is usually thought to be less challenging due to the varying lighting conditions, significant image noise and occlusions in YCB-Video Dataset.

\begin{table*}[htbp]
\begin{center}
\caption{Evaluation results (\small{ADD}-0.1d ACC) on the LineMOD dataset. Symmetric objects are denoted in bold.}
\label{tab:LineMOD-ADD}
\resizebox{0.9\linewidth}{!}{
\begin{tabular}{l|c|c|c|c|c|c|c|c|c|c}
    \toprule[2pt]
 & PoseCNN~\cite{posecnn} & PVNet~\cite{peng2019pvnet} & CDPN~\cite{li2019cdpn} & DOPD~\cite{zakharov2019dpod} & PointFusion~\cite{xu2018pointfusion} & DenseFusion~\cite{densefusion} & G2L-Net~\cite{chen2020g2l} & PVN3D~\cite{pvn3d} & FFB6D~\cite{ffb6d} & Our Uni6D \\
    \midrule[1.5pt]
      ape & 77.0 & 43.6 & 64.4 & 87.7 & 70.4 & 92.3 & 96.8 & 97.3 & 98.4 & 93.71 \\
      benchvise & 97.5 & 99.9 & 97.8 & 98.5 & 80.7 & 93.2 & 96.1 & 99.7 & 100.0 & 99.81 \\
      camera & 93.5 & 86.9 & 91.7 & 96.1 & 60.8 & 94.4 & 98.2 & 99.6 & 99.9 & 95.98\\
      can & 96.5 & 95.5 & 95.9 & 99.7 & 61.1 & 93.1 & 98.0 & 99.5 & 99.8 & 99.02\\
      cat & 82.1 & 79.3 & 83.8 & 94.7 & 79.1 & 96.5 & 99.2 & 99.8 & 99.9 & 	98.10\\
      driller & 95.0 & 96.4 & 96.2 & 98.8 & 47.3 & 87.0 & 99.8 & 99.3 & 100.0 & 99.11\\
      duck & 77.7 & 52.6 & 66.8 & 86.3 & 63.0 & 92.3 & 97.7 & 98.2 & 98.4 & 89.95\\
      \textbf{eggbox} & 97.1 & 99.2 & 99.7 & 99.9 & 99.9 & 99.8 & 100.0 & 99.8 & 100.0 & 100.00 \\
      \textbf{glue} & 99.4 & 95.7 & 99.6 & 96.8 & 99.3 & 100.0 & 100.0 & 100.0 & 100.0 & 99.23\\ 
      holepuncher & 52.8 & 82.0 & 85.8 & 86.9 & 71.8 & 92.1 & 99.0 & 99.9 & 99.8 & 90.20\\
      iron & 98.3 & 98.9 & 97.9 & 100.0 & 83.2 & 97.0 & 99.3 & 99.7 & 99.9 & 99.49\\
      lamp & 97.5 & 99.3 & 97.9 & 96.8 & 62.3 & 95.3 & 99.5 & 99.8 & 99.9 & 99.42\\
      phone & 87.7 & 92.4 & 90.8 & 94.7 & 78.8 & 92.8 & 98.9 & 99.5 & 99.7 &97.41\\
      \hline
        Avg & 88.6 & 86.3 & 89.9 & 95.2 & 73.7 & 94.3 & 98.7 & 99.4 & 99.7 & 97.03 \\   \bottomrule[2pt]
\end{tabular}}
\end{center}                        
\end{table*}

\section{Quantitative Results on the Occlusion LineMOD
dataset}
We follow the previous works~\cite{densefusion,ffb6d} to train our model on the LineMOD dataset and only use this dataset for testing.
Experimental results of LineMOD dataset are reported in Table~\ref{tab:ocLineMOD-ACC}, and we obtain 30.71 ADDS-0.1d AUC.

\begin{table*}[htbp]
\begin{center}
\centering
\caption{Evaluation results (\small{ADD}-0.1d ACC) on the Occlusion-LineMOD dataset. Symmetric objects are denoted in bold.}
\label{tab:ocLineMOD-ACC}
\resizebox{0.9\linewidth}{!}{
\begin{tabular}{l|c|c|c|c|c|c|c|c|c|c|c}
    \toprule[2pt]
 Method & PoseCNN~\cite{posecnn} & Oberweger~\cite{oberweger2018making} & Pix2Pose~\cite{kiru2019pixel} & PVNet~\cite{peng2019pvnet} & DPOD~\cite{zakharov2019dpod} & Hu~\cite{hu2020single} & HybridPose~\cite{song2020hybridpose} & PVN3D~\cite{pvn3d} & FFB6D~\cite{ffb6d} & Our  Uni6D \\
    \midrule[1.5pt]
      ape & 9.6 & 12.1 &  22.0 & 15.8 & - & 19.2 & 20.9 & 33.9 & 47.2 & 32.99\\
      can & 45.2 & 39.9 &  44.7 & 63.3 & - & 65.1 & 75.3 & 88.6 & 85.2 & 51.04\\
      cat & 0.9 & 8.2 &  22.7 & 16.7 & - & 18.9 & 24.9 & 39.1 & 45.7 & 4.56\\
      driller & 41.4 & 45.2 &  44.7 & 65.7 & - & 69.0 & 70.2 & 78.4 & 81.4 & 58.40 \\
      duck & 19.6 & 17.2 &  15.0 & 25.2 & - & 25.3 & 27.9 & 41.9 & 53.9 & 34.80\\
      \textbf{eggbox} & 22.0 & 22.1 &  25.2 & 50.2 & - & 52.0 & 52.4 & 80.9 & 70.2 & 1.73\\
      \textbf{glue} & 38.5 & 35.8  & 32.4 & 49.6 & - & 51.4 & 53.8 & 68.1 & 60.1 & 30.16\\ 
      holepuncher & 22.1 & 36.0 & 49.5 & 39.7 & - & 45.6 & 54.2 & 74.7 & 85.9 & 32.07\\
        Avg & 24.9 & 27.0  & 32.0 & 40.8 & 47.3 & 43.3 & 47.5 & 63.2 & 66.2 & 30.71\\   \bottomrule[2pt]
\end{tabular}}
\end{center}                        
\end{table*}

\section{More Qualitative Results}
\begin{figure*}[tbp]
    \centering
    \includegraphics[width=0.9\textwidth]{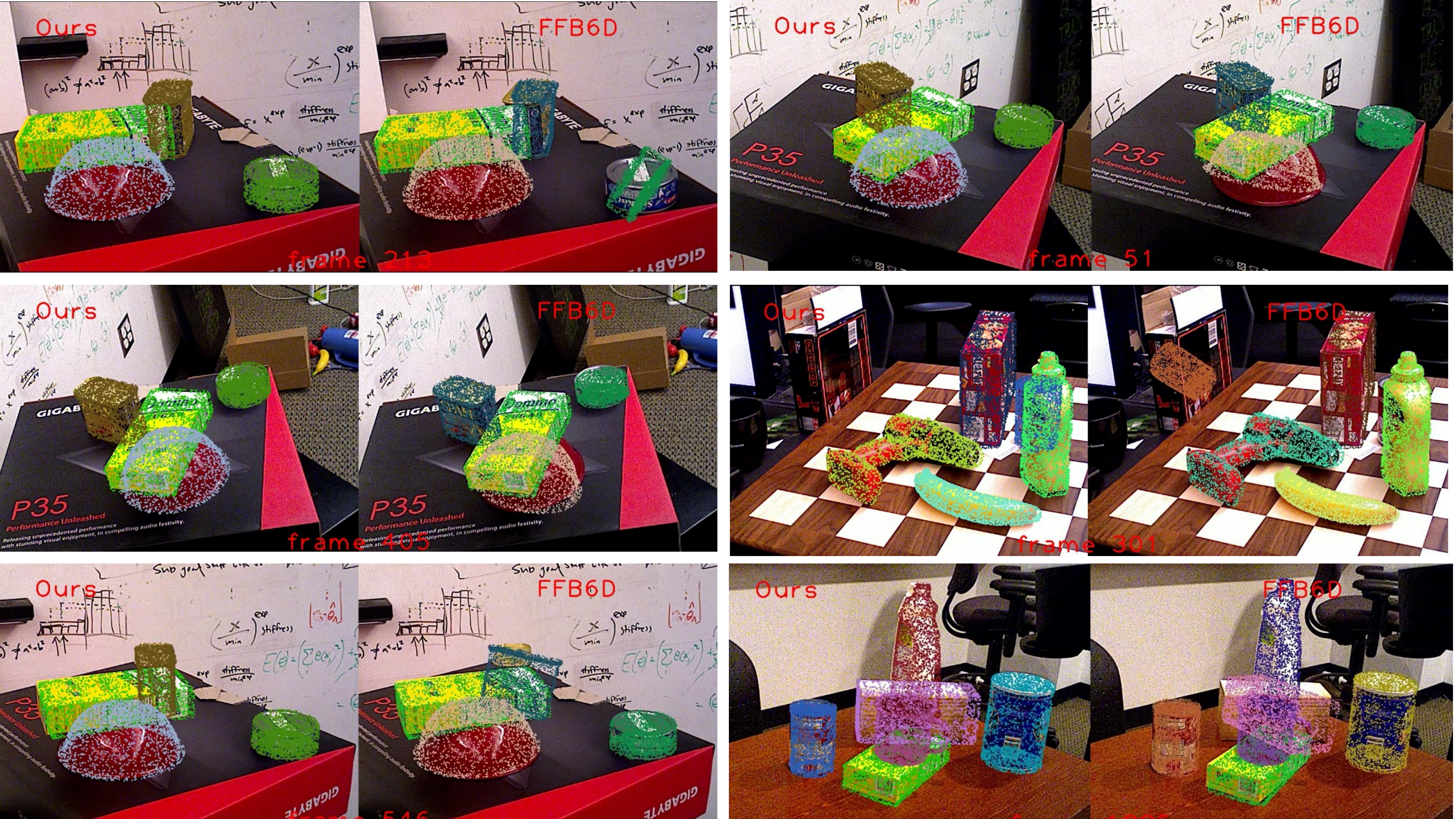}
    \caption{Qualitative results of 6D pose on the YCB-Video dataset. In each sub-figure, left is the result of our method and the right is of the SOTA method FFB6D~\cite{ffb6d}.}
    \label{fig:vis-ycb}
\end{figure*}

\begin{figure*}[th]
    \centering
    \includegraphics[width=0.8\textwidth]{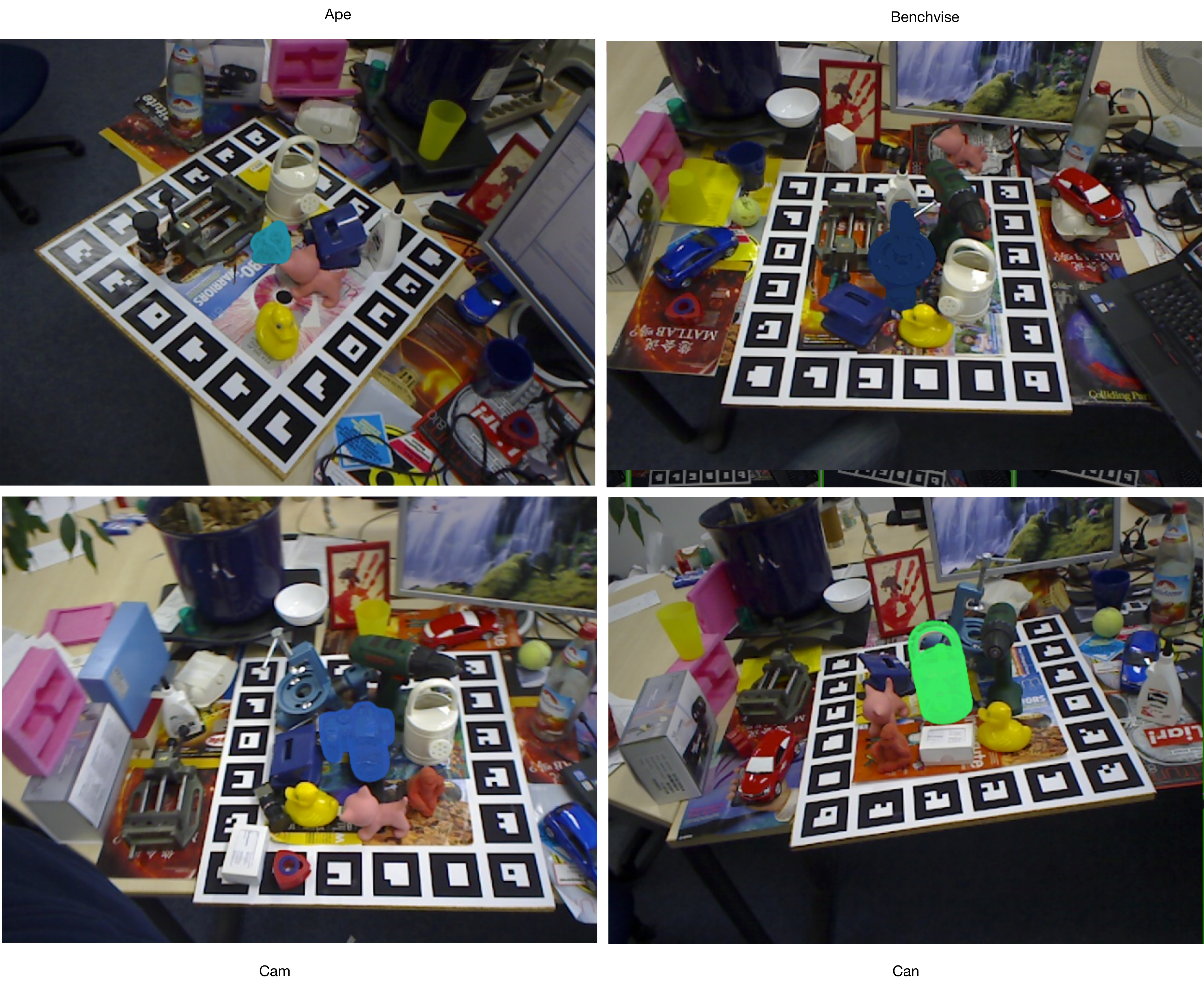}
    \caption{Qualitative results of 6D pose on the LineMOD dataset.}
    \label{fig:vis-lm}
\end{figure*}

We give more qualitative comparison results between our method and the SOTA method FFB6D~\cite{ffb6d} in Fig.~\ref{fig:vis-ycb} for YCB-Video dataset and Fig.~\ref{fig:vis-lm} for LineMOD dataset. 
Moreover, \textbf{We strongly recommend readers to watch the video from \href{https://youtu.be/6G__P282djw}{https://youtu.be/6G\_\_P282djw} , which directly reflects the comparison results between our method and the FFB6D~\cite{ffb6d}.}
Compared with FFB6D, our method estimates the 6d pose more smoothly. Our method has better consistency between adjacent frames, less jitter, and more robust performance under severe occlusion conditions. 

